\documentclass[conference]{IEEEtran}
\ifCLASSINFOpdf
\else
\fi
\usepackage{graphicx}
\usepackage{amsmath}
\usepackage{filecontents}
\usepackage{caption}
\usepackage{float}
\usepackage{subcaption}
\usepackage{color,soul}
\usepackage{multirow}
\usepackage[ruled,vlined,linesnumbered]{algorithm2e}
\usepackage{url}
\usepackage[table]{xcolor}
\usepackage{booktabs,pdflscape,adjustbox}

\begin{document}
\title{Deep Reinforcement Learning: Framework, Applications, and Embedded Implementations\\
{\large Invited Paper}}

\author{Hongjia Li$^1$, Tianshu Wei$^2$, Ao Ren$^1$, Qi Zhu$^2$, and Yanzhi Wang$^1$ \\
\normalsize{$^1$Dept. Electrical Engineering \& Computer Science, Syracuse University, Syracuse, NY, USA}\\
\normalsize{$^2$Dept. Electrical \& Computer Engineering,
University of California, Riverside, CA, USA} \\
\normalsize{$^1$\{hli42, aren, ywang393\}@syr.edu}, $^2$\{twei002@ucr.edu, qzhu@ece.ucr.edu\}}

\maketitle

\begin{abstract}
The recent breakthroughs of deep reinforcement learning (DRL) technique in Alpha Go and playing Atari have set a good example in handling large state and actions spaces of complicated control problems. The DRL technique is comprised of (i) an offline deep neural network (DNN) construction phase, which derives the correlation between each state-action pair of the system and its value function, and (ii) an online deep Q-learning phase, which adaptively derives the optimal action and updates value estimates.

In this paper, we first present the general DRL framework, which can be widely utilized in many applications with different optimization objectives. This is followed by the introduction of three specific applications: the cloud computing resource allocation problem, the residential smart grid task scheduling problem, and building HVAC system optimal control problem. The effectiveness of the DRL technique in these three cyber-physical applications have been validated.
Finally, this paper investigates the stochastic computing-based hardware implementations of the DRL framework, which consumes a significant improvement in area efficiency and power consumption compared with binary-based implementation counterparts.
\end{abstract}


\section{Introduction}

Reinforcement learning provides us a mathematical framework for learning or deriving strategies or policies that map situations (i.e., states) into actions with the goal of maximizing an accumulative reward \cite{sutton1998reinforcement}. It has been widely applied for solving problems in different fields, such as manufacturing, finance sector, and robotic control systems. Along with the resurgence of deep learning techniques, reinforcement learning has now evolved towards deep reinforcement learning (DRL), where deep neural networks (DNNs) are utilitzed in the policy-deriving process \cite{mnih2013playing,mnih2015human,silver2016mastering}. With offline-constructed and online-updated DNNs, DRL techniques demonstrate capabilities in handling complicated problems with high-dimensional state and action spaces and even enabling continuous action spaces 
\cite{lillicrap2015continuous}. These features make DRL distinguished from reinforcement learning. And recent breakthroughs in Alpha Go \cite{silver2016mastering} and playing Atari \cite{mnih2013playing} indicate the great success of DRL.

One major application scenario of DRL is the embedded computing environment, such as in unmanned aerial vehicles, autonomous driving, robotics, wearable devices and mobile computing systems. However, DNNs involved in the DRL can be both compute and memory intensive. Therefore, it is desirable to have dedicated hardware implementations (e.g., FPGA, ASIC) for DNNs in the DRL for the embedded computing platforms, in order to utilize the distributed-computing and parallelism of hardware resources for enhanced computing speed, energy efficiency, and resiliency. Stochastic computing (SC) \cite{alaghi2013survey,gaines1969stochastic} as a low-cost substitute to the binary-based computing radically simplifies the hardware implementation of arithmetic units and has the potential to satisfy the low power and small hardware footprint requirements of DNNs in the embedded computing environment.

In this paper, we first present the general DRL framework, which can be widely utilized in many applications with different optimization objectives, such as resource allocation, residential smart grid, embedded system power management, and autonomous control. Followed by the introduction of three applications of the DRL framework, one for the cloud computing resource allocation problem, one for the residential smart grid user-end task scheduling problem and one for building HVAC system. The cloud computing resource allocation problem automatically and dynamically distributes resources (virtual machines or tasks) to servers by establishing efficient strategy. Through extensive experimental simulations using Google cluster traces \cite{clusterdata:Reiss2011}, the DRL framework for cloud computing resource allocation achieves up to 54.1\% energy saving compared with the baseline approach. The residential smart grid task scheduling problem determines the task scheduling and resource allocation with the goal of simultaneously maximizing the utilization of photovoltaic (PV) power generation and minimizing user's electricity cost. Through extensive experimental simulations with realistic task modelings, the DRL framework for residential smart grid task scheduling achieves up to 22.77\% total energy cost reduction compared with the baseline algorithm. The building HVAC system is designed for controlling a desired temperature within each zone with the factors of current zone temperature and outside environment disturbances. The proposed DRL control algorithm can achieve 20\%-70\% cost reduction compared with the rule-based baseline control strategy, while maintaining the temperature violation rate below 1.0\%. 


Additionally, as mentioned above, this paper investigates the stochatic computing (SC)-based hardware implementations of DNNs used in DRL using stochastic computing technique. To further enhance the performance (computing speed) and energy efficiency, pipelining techniques is employed in the SC-based hardware design. The stochastic computing-based ultra-low-power implementation consumes only 58771.53 $\mu m^{2}$ area and 7.73 $mW$ power with 261.12 $ns$ delay.

The rest of this paper is organized as as follows. Section 2 presents the related works on DRL. In Section 3, the general DRL basics and framework are introduced. Section 4 introduces three representative applications of DRL, along with simulation results. In the following Section 5, the hardware implementation of DRL using the stochastic computing technique is presented. The corresponding experimental results are showed in Section 6. The conclusion of this paper is presented in Section 7.  

\section{Related Works}
A lot of research efforts have been made recently on the development and applications of DRL. Mnih et al. are the first introducing deep learning model into the reinforcement learning and have succeeded in handling high-dimensional sensory input when playing Atari \cite{mnih2013playing}. 

In 2015, Mnih et al. further generalized DRL by developing the first artificial agent, called deep Q-network (DQN), capable of learning policies directly from high-dimensional sensory inputs and agent-environment interactions \cite{mnih2015human}, in which convolutional neural networks with hierarchical layers of tiled convolutional filters were adopted. 
Lillicrap et al. proposed an actor-critic, model-free algorithm based on the deterministic policy gradient. Combined with DQN, the actor-critic approach can operate over continuous action spaces \cite{lillicrap2015continuous}.
In 2016, Silver et al. combined supervised learning from games of human experts and reinforcement learning from self-play games to master the game of Go with DNN and tree search \cite{silver2016mastering}. In \cite{van2016deep} a specific adaptation to the DQN algorithm with double Q-learning was proposed, which is able to reduce the observed overestimations of the original DQN algorithm, and also lead to much better performance on several games including the Atari 2600 domain. 

There are also extensive research works on enhancing the performance and energy efficiency of hardware implementations of DNNs. 
In order to effectively implement the deep convolutional neural networks onto embedded/portable systems, Ren et al. developed the first comprehensive design and optimization framework of stochastic computing-based deep convolutional neural networks \cite{ren2016sc}. In order to handle the challenges brought by stochastic computing including random error fluctuation, range limitation, and overhead in accumulation, Kim et al. adopted the approach of removing near-zero weights, applying weight-scaling, and integrating the activation function with the accumulator when designing an efficient DNN with stochastic computing \cite{kim2016dynamic}. In \cite{shafiee2016isaac}, a pipelined architecture was employed for a convolutional neural network accelerator, with memristor crossbars dedicated for each neural network layer and eDRAM data buffers between pipeline stages. 
Ardakani et al. implemented the DNN using integer stochastic stream which is a sequence of integer numbers that are represented by either two's complement or sign-magnitude \cite{ardakani2017vlsi} to solve the precision loss issue of conventional scaled adder, meanwhile reducing the latency.

\section{DRL Framework}
Deep reinforcement learning shares the same basic concepts with reinforcement learning in that it is also an agent-environment interaction process. The learner and decision-maker is called the \emph{agent}. The thing it interacts with, comprising everyting outside the agent, is called the \emph{environment}. Specifically, the agent and environment interact at a sequence of decision epochs. At a decision epoch, the agent receives some representation of the environment's \emph{state} i.e., $s$, and on that basis selects an \emph{action} i.e., $a$. In part as a consequence of its action, the agent receives a numerical \emph{reward} and finds itself in a new state of the environment i.e., $s'$. A policy, denoted by $\pi$, of the agent is a mapping from each state to an action that specifies the action $a=\pi(s)$ that the agent will choose when the environment is in state $s$. The ultimate goal of an agent is to find the optimal policy, such that
\begin{equation}\label{eqn_1}
V^{\pi}(s) = \mathbf{E} \Big[\sum_{k=0}^{\infty }\gamma^kr(k) \Big| s\Big]
\end{equation}
or
\begin{equation}\label{eqn_2}
V^{\pi}(s)  = \mathbf{E} \Big[\int_{t_0}^{\infty}e^{-\beta(t-t_0)}r(t) dt \big| s\Big]
\end{equation}
is maximized for each state $s$, where $r$ is the reward rate, and $\gamma$ and $\beta$ are the discount rates. The \emph{value function} $V^{\pi}(s)$ is the expected return when the environment starts in state $s$ and follows policy $\pi$ thereafter. Eqn. (\ref{eqn_1}) is for a discrete-time system, while Eqn. (\ref{eqn_2}) is for a continuous-time system. 

In order to derive the optimal policy, a $Q$ value, denoted by $Q(s,a)$, is associated with each state-action pair $(s,a)$, which approximates the expected discounted cummulative reward (i.e., the value function) of taking action $a$ at state $s$. The reinforcement learning algorithm has a convergence time proportional to $O(|A| \cdot |S|)$, where $|A|$ represents the total number of actions and $|S|$ represents the total number of states. And its computation complexity is $O(|A|+M)$ at each decision epoch, in which $M$ is the already known state-action pairs kept in the memory. Therefore, reinforcement learning becomes less effective when dealing with actual complicated problems with high-dimensional state and action spaces.     

To overcome the drawbacks of reinforcement learning, DRL is comprised of an offline deep neural network (DNN) construction phase and an online deep Q-learning phase showed in Algorithm \ref{algorithm1}. In the offline phase, we construct a DNN, which can infer for each state-action pair its $Q$ value to be used for the online phase. Sufficient training data is needed for the offline DNN construction. In \cite{rao2009vconf} a model-based procedure is adopted to accumulate the training samples, while in \cite{silver2016mastering} training data is obtained from actual measurement. To obtain the training data, we use an arbitrary but gradually refined policy to simulate the control process. An experience memory $D$ with capacity $N_{D}$ is used to store the state transition profiles and $Q$ values while smoothing out learning to avoid oscillations and divergence in the parameters \cite{mnih2013playing}. Then, a DNN with weight set $\theta$ can be trained using the state transition profile and $Q$ values. 

\begin{algorithm}[t]
\caption{The General DRL Framework}\label{algorithm1}
\textbf{Offline DNN construction:}\\
\nl Simulate the control process using an arbitrary but gradually refined policy for enough long time\;
\nl Obtain the state transition profile and $Q(s,a)$ value estimates during the process simulation\;
\nl Store the state transition profile and $Q(s,a)$ value estimates in experience memory $D$ with capacity $N_D$\;
\nl Train a DNN with features $(s,a)$ and outcomes $Q(s,a)$\;
\textbf{Online deep Q-learning:}\\
\nl \ForEach{execution sequence}
{
	\nl \ForEach{decision epoch $t_k$}{
		\nl With probability 1 - $\varepsilon$ select the action $ a_k  = argmax_a Q(s_k,a)$, otherwise randomly select an action\;
		\nl Execute the chosen action in the control system\;
		\nl Observe reward $r_k(s_k,a_k)$ during time period $[t_k,t_{k+1})$ and the new state $s_{k+1}$ at the next decision epoch\;   
		\nl Store transition set $(s_k,a_k,r_k,s_{k+1})$ in $D$\;
		\nl Update $Q(s_k,a_k)$ based on $r_k$ and $max_{a^{'}} Q(s_{k+1},a^{'})$ based on Q-learning updating rule. One could use a duplicate DNN $\hat{Q}$ to achieve this goal\;
		}
	\nl Update DNN weight set $\theta$ based on updated Q-value estimates, in a mini-batch manner\;
}

\end{algorithm}

In the online phase, deep Q-learning is adopted for action selection (i.e., the $\varepsilon$-greedy policy) and $Q$ value update. Specifically, suppose at decision epoch $t_k$, the system under control is in state $s_k$. The DRL agent enumerates all actions and obtains the corresponing $Q(s_k,a)$ value estimates using the offline-constructed DNN. According to the $\varepsilon$-greedy policy, the agent selects the action resulting in the maximum $Q(s_k,a)$ value estimate with probability 1 - $\varepsilon$, and selects a random action with probability $\varepsilon$. After the selected action $a_k$ is taken, the observed total reward $r_k(s_k,a_k)$ during $[t_k,t_{k+1})$ is used for $Q$ value update. 
In order to mitigate the potential oscillation in the DNN inference results, we adopt the duplicate $Q$ method from \cite{hasselt2010double}, which maintains two $Q$ value estimates for each state-action pair and updates the two $Q$ value estimates interactively. At the end of an execution sequence of decision epochs, the DNN is then updated using the lately observed $Q$ values in a mini-batch manner, and will be employed in the next execution sequence. 

From the above procedure, the DRL can now handle extremely large state space (even infinite continuous state space) by using offline-trained and online-updated DNN. For the action space, it should be kept within a reasonable size, due to the necessity to enumerate the action space for action selection at a decision epoch.

\section{Representative Applications of Deep Reinforcement Learning}
\subsection{DRL Framework for Cloud Computing Resource Allocation}



In the cloud computing resource allocation problem, a server cluster consists of $M$ physical servers that can provide $P$ types of resources is considered. A first-come-first-served manner is deployed to process assigned jobs for the servers. A job will wait in the queue until sufficient resource is released in the server. We define the latency of a job as the actual duration from its arrival time to its complete time. 

A server has two working modes: active and sleep for energy saving. $T_{on}$ is the time needed by a server to transit from sleep mode to active mode.  $T_{off}$ is the time needed by a server to transit from active mode to sleep mode when no job is pending or running. All the mode transitions are considered as uninterruptible. We assume the power consumption of a server in the sleep mode is zero. Based on an empirical non-linear model in \cite{fan2007power}, the power consumption of a server in active mode is a function of CPU utilization as follows: 
\begin{equation}
P(u_t) = P(0\%)+(P(100\%)-P(0\%))(2u_t-u_t^{1.4})
\end{equation}
where $u_t$ denotes the CPU utilization of the server at time $t$.

In order to significantly reduce the action space, we adopt a continuous-time and event-driven decision making mechanism \cite{duff1995reinforcement} in which each decision epoch coincides with the arrival time of a new job.
In the offline phase, we harness the power of \emph{representation learning} and \emph{weight sharing} for DNN construction. Specifically, we first employ an autoencoder to extract a lower-dimensional high-level representation of server group state for each possible server. The dimension difference reflects the relative importance of the targeting server group compared with other groups and results in reduction in the state space. Next, for estimating the Q-value of the action of allocating a job to servers in this group the neural network $\textbf{Sub-Q}$ takes the server group state, job's state, all lower-dimensional high-level representations, and actions as input features. In addition, we introduce weight sharing among all autoencoders, as well as all $\textbf{Sub-Q}$'s to reduce the total number of parameters and the training time. For the online phase, at the beginning of each decision epoch, the Q value estimates are derived for each state-action pair by inference based on the offline trained DNN. An action is then selected for the current state using the $\epsilon$-greedy policy. At the next decision epoch, Q-value estimates are updated. After the execution of a whole control procedure, the DNN is updated in a mini-batch manner with the newly observed Q-value estimates.

In the simulation setup, we assume a homogeneous server cluster without loss of generality. The idle power consumption is $P(0\%) = 87$W, and the peak power consumption is $P(100\%) = 145$W \cite{fan2007power}. 
We set the server power mode transition times $T_{on}=30$s and $T_{off}=30$s. 
Based on the Google cluster traces \cite{clusterdata:Reiss2011}, we simulate five different one-week job traces into the proposed online deep Q-learning framework and compare the average results against the baseline. Under the circumstances of $M$ = 20, 30 and 40, the proposed DRL-based framework on average can achieve 20.3\%, 47.4\% and 54.1\% of power consumption saving while the accumulated latency only increases by 9.5\%, 16.1\% and 18.7\%. 
The proposed framework effectively generates policies to decrease accumulated latency when the weight increases because of the more evenly jobs distributing. All tested cases can achieve at least 47.8\% power consumption saving with only a slight increase in job latency. These results prove that weights of the reward function can take a effective control of the trade-off between power, latency, and resiliency.

\subsection{Residential Smart Grid Task Scheduling}
The present research focuses on task scheduling of residential appliance operations to minimize an individual electricity user's cost in the Smart Grid factoring in photovoltaic (PV) power generation, due to the worldwide trend of transition to the Smart Grid and PV power usage in residential, industrial, and commercial sectors. In this work, we reduce users' electricity cost by applying the deep reinforcement learning framework for the user-end task scheduling in the Smart Grid equipped with distributed PV power generation devices under dynamic pricing.


We employ a slotted time model i.e., the task scheduling frame (one day) is divided into $T = 24$ time slots each with duration of one hour. The tasks are non-interruptible, i.e., tasks need to be operated in continuous time slots. An inconvenience price is determined by the user to represent the penalty when scheduling task outside its desired operating window.
We assume that the residential user is equipped with a distributed PV system.
The power generation of the PV system in time slot $t$ is denoted by $P_{pv}(t)$.
The power provided from the grid in time slot $t$ is denoted as $P_{grid}(t)$, which depends on $P_{pv}(t)$ and $P_{load}(t)$ according to the following:
\begin{equation}
P_{grid}(t) =
\begin{cases}
0, & \text{when}\ P_{pv}(t) \geq P_{load}(t)\\
P_{load}(t)-P_{pv}(t), & \text{otherwise}
\end{cases}
\end{equation}
We consider a dynamic price model $C(t,P_{grid}(t))$ consisting of a time-of-use (TOU) price component and a power consumption price component. 

We simulate the control process using generated task sets and following a preliminary control policy. 
The state transition profile and $Q(s,a)$ value estimates are obtained through the simulation and used as the training data for offline DNN construction.
We construct a three-layer artificial neural network with 26 hidden neurons, which is trained using the previously obtained training data.
In the online phase, for each decision epoch $k$, according to the current system state $s_k$, the action resulting in the maximum $Q(s_k,a)$ estimate is selected using the $\epsilon$-greedy policy.
And $Q(s_k,a)$ estimates are obtained by performing inference on the offline-trained neural network.
Based on the selected actions and observed rewards, Q-value estimates are updated before the next decision epoch.
At the end of one execution sequence, the neural network is updated for use in the next execution sequence.



The PV power generation profiles are provide by \cite{PVpower}, which are measured at Duffield, VA, in 2007. We adopt an approach using the negotiation-based task scheduling algorithm \cite{li2014negotiation} as our baseline system. We compare the total electric cost for the residential smart grid user using the DRL framework and the baseline algorithm on the following test cases: 100, 300 and 500 tasks for scheduling.  
According to the results, the DRL framework can schedule tasks to maximize the coverage of the PV power and avoid the peak of TOU price in a more effective manner compared to the baseline method. Correspondingly, the DRL framework can achieve $22.77\%$, $12.54\%$ and $12.45\%$ total energy cost reductions when the number of tasks are 100, 300, and 500, respectively.

\subsection{DRL for Building HVAC Control}
The building HVAC system should be operated to maintain a desired temperature within each zone, based on current zone temperature and outside environment disturbances (e.g., ambient temperature and solar irradiance). The zone temperature at next time step is determined by the current system states, the environment disturbances, and the conditioned air input from the HVAC system. We have developed a DRL control algorithm to intelligently determine the optimal conditioned air flow input for each zone, for maintaining desired temperature while minimizing the total energy cost of the building HVAC system~\cite{Wei_DAC17}.

More specifically, we consider a building that is equipped with a VAV (variable air flow volume) HVAC system to maintain desired temperature for $z$ zones. The VAV terminal box in each zone provides conditioned air (typically at a constant temperature) with an air flow rate that can be chosen from multiple discrete levels (denoted as $F=\{f^1,f^2,...,f^m\}$). At each control time step, the optimal control action for each zone is determined based on the observation of the current system states, which include current physical time, zones' temperature in the building and environment disturbances (i.e. ambient temperature and solar irradiance intensity). For environment disturbances, we also take into account a multi-step forecast of weather data in the system states. This enables our DRL algorithm to capture the trend of the weather condition and perform proactive control for time-variant systems. 

We separately train a neural network for each zone by following the DRL Algorithm~\ref{algorithm1}. Each neural network is only responsible for approximating the Q-value in one zone. At each control time step, all neural networks will receive the entire system states of buildings and then determine the control action for each zone separately. This heuristic can greatly improve the training efficiency by reducing the number of output units in the neural network. 
\begin{align}
&r^i_{t} =  -\lambda ( [{T_{t}^i} - {\overline{T}_{t}^i}]_+ + [{\underline{T}_{t}^i} -  {T_{t}^i}]_+ ) -\hspace{30mm}\nonumber\\
&\hspace{8mm}cost(\sum_i a^i_{t-1}, s_{t-1})\cdot\frac{a^i_t}{\sum_i a^i_{t-1}},\hspace{3mm}a^i_{t-1}\in F\label{rewards}
\end{align}

During the training process, our DRL algorithm will try to maximum the reward function~\eqref{rewards} for each zone. The first term measures the temperature violation in each zone, while the second term heuristically estimates the energy consumption cost contributed by each zone (which is assumed to be proportional to the air flow demand in each zone based on the total HVAC system energy cost in the building $cost(\cdot)$).

To calculate the Q-value estimates, we adopt a similar neural network structure as in~\cite{mnih2013playing}. Each output unit in the neural network corresponds to the Q-value estimate of each available control action. By using this structure, the Q-value estimates for all control actions can be calculated by performing one forward pass. We calculate the optimal Q-value for the action in the current system state by following the Bellman Equation~\eqref{target}.
\begin{align}
Q^*(s_{t-1},a_{t-1}) &= r_{t}+\gamma\max_{a_{t}}Q(s_{t},a_{t})\label{target}\\
&\Leftarrow\max[\frac{r_{t}}{\rho}+\gamma\underset{a_{t}}{\max}Q(s_{t},a_{t}), -1]\label{clipping}
\end{align}

As shown in Equation~\eqref{clipping}, in practice we squash the original target Q-value to the range $[-1,0]$ by first shrinking the original reward with a factor $\rho$ and then clipping it if the target Q-value estimate is smaller than $-1$. This can help speedup the training process by reducing the variance of Q-value estimates.

We train the DRL algorithms on two different weather profiles in summer days. The first set of weather data has intensive solar radiation and large variance in temperature, while the second one has a milder weather profile. We calculate buildings' energy cost by using the practical time-of-use price from the Southern California Edison, and demonstrate the effectiveness of our DRL algorithm by comparing with a rule-based HVAC control strategy (similarly as the one in~\cite{Urieli:AAMAS2013}) and the conventional RL method. We evaluate the performance of our DRL algorithm with three building models, which have 1 zone, 4 zones and 5 zones, respectively. Our experiment results show that our DRL control algorithm is superior to the conventional RL method and is able to achieve $20\%-70\%$ cost reduction compared with the rule-based baseline control strategy, while maintaining the temperature violation rate below $1.0\%$~\cite{Wei_DAC17}.

\section{SC-Based DRL Implementation}

Compared with conventional implementations in CMOS circuits, stochastic computing (SC) enables low-power and small-hardware-footprint implementations of arithmetic units using standard logic elements \cite{gaines1967stochastic}. The SC paradigm significantly simplifies the hardware implementation and thereby allowing very high clock rates.
In addition, it can provide a high degree of fault tolerance and an opportunity for trade-off between computating speed and accuracy even without changing the hardware implementation.  

In stochastic computing (SC), bit-streams are used for representing numbers. First, the occurance rate of 1's i.e., $P(X=1)$ in a bit-stream is calculated. Next, according to unipolar encoding the number $x$ presented by the bit-stream is just $x=P(X=1)$, or according to bipolar encoding the number $x$ presented by the bit-stream is $x=2P(X=1)-1$ \cite{brown2001stochastic}. A bit-stream can represnt a number in the range of $[0, 1]$ in unipolar encoding or $[-1,1]$ in bipolar encoding. For representing a number beyond the range, a pre-scaling operation \cite{yuan2016design} is needed. In this paper, we choose bipolar encoding to cover both negtive and positive numbers in DNN related calculations. For instance, a bit-stream 1101001011 represents the number $0.2$.

\subsection{SC Arithmetic Units}

The major arithmetic operations in DNNs are multiplication, addition, and activation function. These operations can be implemented with extremely small arithmetic units as follows.

\begin{figure}
\centering
\includegraphics[width=3.2in]{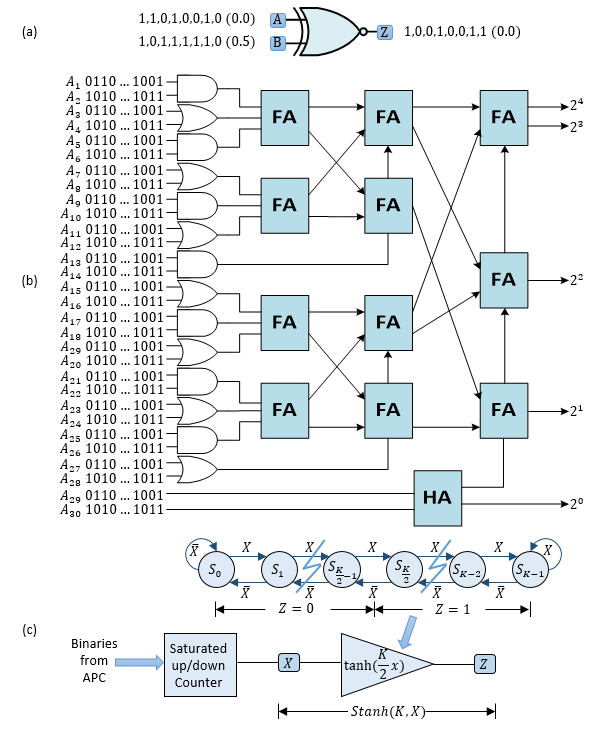}
\caption{SC arithmetic units used in this work. (a) XNOR gate-based mulipication unit, (b) APC-based addition unit performing addition of 30 bit-streams, and (c) $K$-state FSM-based activation unit.}\label{fig:AU}
\end{figure}






\textbf{Multiplication Unit:} The multiplication of two numbers represented by bit-streams (in bipolar encoding) can be calculated as logic XNOR operation of the two bit-streams, as shown in Figure \ref{fig:AU} (a). A brief derivation can be $a\times b = [2P(A=1)-1]\times[2P(B=1)-1]=2P(A=1)P(B=1)+2P(A=0)P(B=0)-1=2P(A=1)\odot P(B=1)-1=2P(Z=1)-1=z$. Regardless of the length of bit-streams (i.e., precision), the multiplication unit is simply an XNOR gate with two 1-bit inputs and one 1-bit output \cite{gaines1969stochastic}.



\begin{figure}
\centering
\includegraphics[width=3.2in]{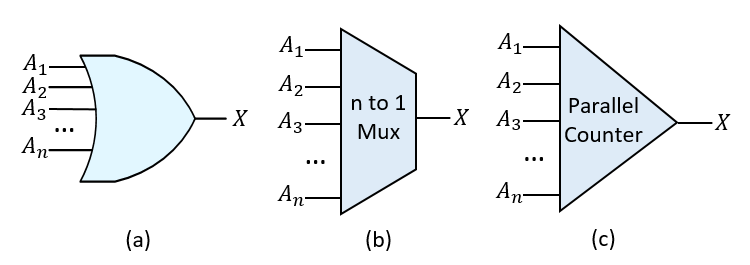}
\caption{Addition units: (a) OR gate, (b) MUX, and (c) APC.}
\label{fig:add}
\end{figure}

\textbf{Addition Unit:} The addition of $n$ numbers can be performed as logic OR operation of the $n$ bit-streams, or by an $n$-to-1 multiplexer where $n$ inputs take the bit-streams respectively and the output bit-stream equals to $1/n$ of the sum, or by an approximate parallel counter (APC) \cite{kim2015approximate}, as shown in Figure \ref{fig:add}, where each of the inputs is a bit-stream. The APC counts the number of ones from its $n$ inputs, that is to say, it adds the $i$-th bit of each of the bit-streams into a $\log n$-bit binary number with the value approximately equivalent to the sum. In summary, OR gate is the most area efficient but the accuracy is too low, MUX is area efficient with limited accuracy, and APC achieves the highest accuracy at the cost of a larger footprint. We adopt APC for addition considering accuracy, power consumption, and footprint according to \cite{ren2016sc}.

An APC employs two parts, an approximate unit (AU), consisting of AND and OR gates for accumulating approximation, and an adder tree consisting of adders to calculate the binary summation of all input bits, each coming from an input bit-stream. We propose an improved APC design as shown in Figure \ref{fig:AU} (b), where the last pair of inputs are feeded to a half adder directly instead of an AND or OR gate. For an APC with 30 inputs as in Figure \ref{fig:AU} (b), the output should be 5-bit binary numbers. 
In order to further reduce the hardware footprint, we employ inverse mirror full adders as proposed in \cite{kim2015approximate} for the adder tree in an APC. Inverse mirror full adders are smaller and more responsive adders that output inverse logic of true summation and carry-out bits. The internal results in the even layers correspond to the number of ones in the primary input, while the internal results in the odd layers represent the number of zeros.
We compare inaccuracy rates of our improved APC design to those of the original APC. As shown in Table \ref{table1:Inaccuracies}, our improved designs significantly reduce inaccuracy rate to less than 0.7\% and at the same time with more than 40\% reduction of gate count.

\begin{table}
\centering
\caption{Inaccuracy rates of the improved and orginal APC designs.}
\label{table1:Inaccuracies}
\renewcommand{\arraystretch}{1.4}
\begin{tabular}{|c|ccc|}
\hline
\multirow{2}{*}{\textbf{APC}} & \multicolumn{3}{c|}{\textbf{Bit Stream Length}} \\ \cline{2-4} 
 & \textbf{256} & \textbf{512} & \textbf{1024} \\ \hline
\textbf{26-input} & 2.56\% & 2.12\% & 1.71\% \\
\textbf{30-input} & 2.34\% & 2.03\% & 1.56\% \\
\textbf{26-input improved} & 0.63\% & 0.61\% & 0.57\% \\
\textbf{30-input improved} & 0.61\% & 0.58\% & 0.55\% \\ \hline
\end{tabular}
\end{table}

\textbf{Activation Unit:} 
The most popular activation functions used for deep neural networks are sigmoid, tanh, and Rectified Linear Unit (ReLU). In this work, we select tanh due to its convenience for SC implementation and comparable effectiveness as ReLU and sigmoid \cite{krizhevsky2012imagenet}.
The tanh function can be easily implemented with a $K$-state finite-state-machine (FSM) in the SC domain with significantly reduced hardware footprint compared to its conventional computing counterpart \cite{brown2001stochastic}.
Figure \ref{fig:AU} (c) includes a $K$-state FSM design of the tanh function in SC domain for use in the activation unit. 
It outputs a zero if the current state is on the left half of the states, and a one otherwise.
By this design, we have
\begin{equation}
Stanh(K,X) \cong tanh(\frac{K}{2}x)
\end{equation} 
where $Stanh$ stands for the tanh function in SC domain.
The $K$ value represents the precision of $Stanh$, and therefore higher accuracy can be achieved with a larger $K$ value. We use a $K$ value in the range of $[-\frac{K}{2}x,\frac{K}{2}x]$ in our experiments. $Stanh(K,X)$ takes bit-streams as input, while inner products calculated from an APC are in the binary format. Therefore, we use a saturated up/down counter \cite{kim2016dynamic} to convert the binary format input from APC to a bit-stream. The whole design of the activation unit is shown in Figure \ref{fig:AU} (c).


\subsection{System Design}

\begin{figure}[t]
\centering
\includegraphics[width=3.2in]{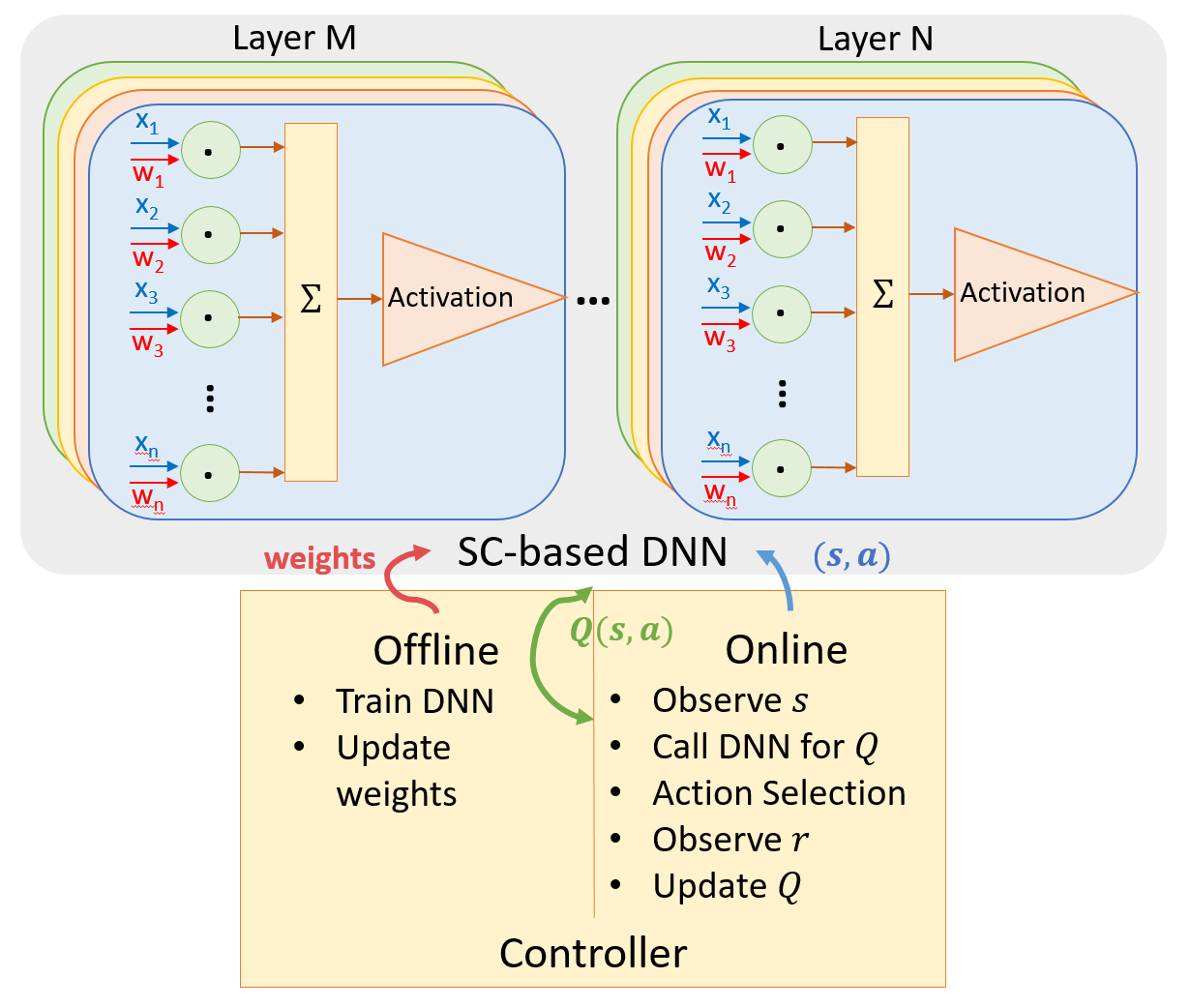}
\caption{Whole system diagram of the DRL implementation including the SC-based hardware DNN and interface with the software controller in an embedded processor. }
\label{fig:overview}
\end{figure}

Figure \ref{fig:overview} shows the whole system diagram of the proposed DRL implementation for embedded computing platforms. It consists of an SC-based hardware DNN, a software controller in an embedded processor, and a B/S conversion block in between, which converts data in binary format for software controller to/from bit-streams for SC-based hardware DNN. For the SC-based hardware DNN, the previously discussed SC arithmetic units including multiplication units, addition units and activation units are utilized to perform DNN calculations. More specifically, the DNN consists of $M$ layers, each with $N_i$ $(1\leq i \leq M)$ neurons. 
The inputs($x_i$) and its corresponding weights($w_i$) are operated by the multiplication and addition units. In order to insure the next layer's input are within [-1,1] range, the outputs are transformed by an activation function.

The software controller performs both online control for each decision epoch and offline control for a sequence of decision epochs. The offline control first constructs a DNN using previously collected data and the resultant weights of the DNN are sent to the hardware DNN as parameters for online inference. The online control at each decision epoch $k$ performs action selection and $Q$ value update, during which state-action pairs $(s_k,a)$ for each action $a$ are sent to the hardware DNN for the calculation of $Q$ values $Q(s_k,a)$ (i.e., DNN inference). $Q$ values calculated from the hardware DNN are then sent back to the software controller for use in action selection and $Q$ value update. After the online execution at a sequence of decision epochs, the offline control takes charge again to update DNN weights with training based on the newly updated $Q$ values.    


\subsection{Design Optimization}

\begin{figure}[t]
\centering
\includegraphics[width=3.2in]{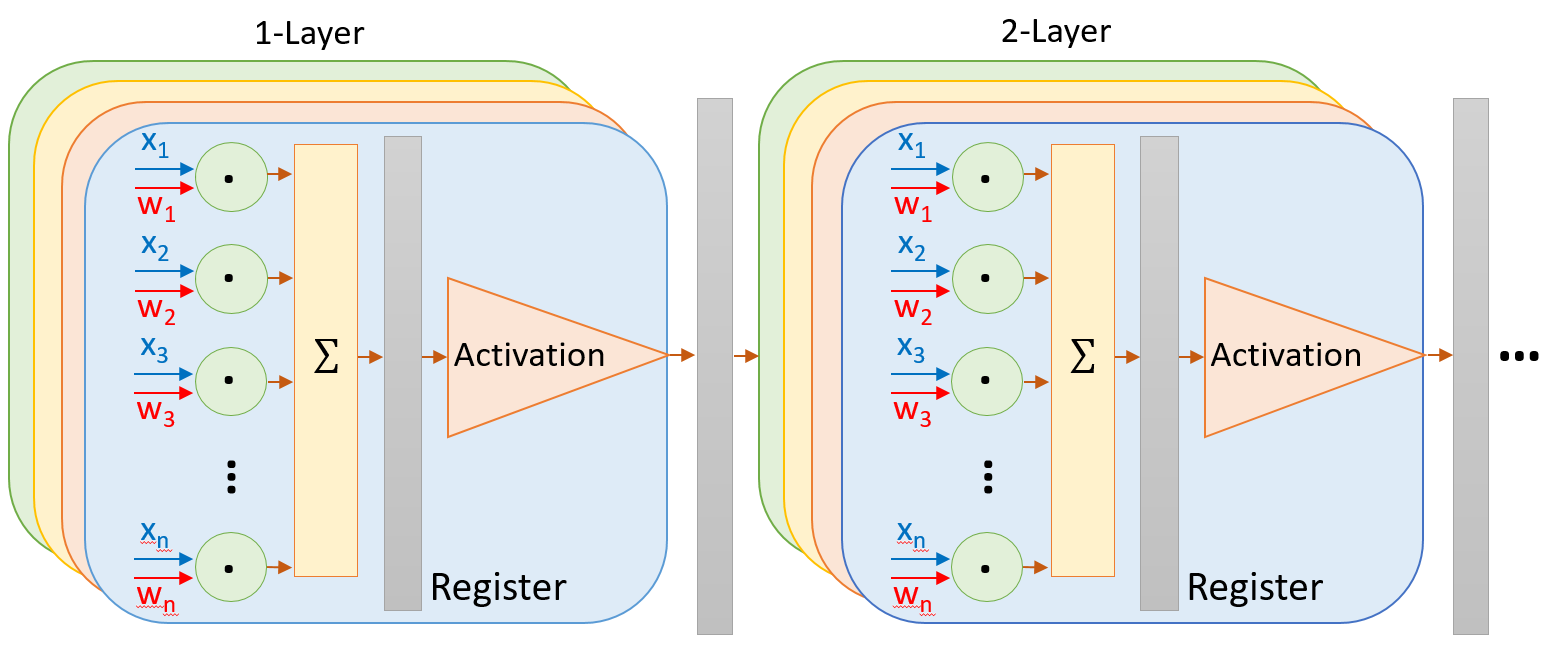}
\caption{Deep pipelining technique in the SC-based hardware DNN.} 
\label{fig:pipeline}
\end{figure}

Different from \cite{ren2016sc}, we use the ``deep'' pipelining technique in the SC-based hardware DNN, where the pipeline stages can be within the DNN layers, while in \cite{ren2016sc} only inter-layer pipelining is considered.
The clock rate of a pipelined architecture is in general increased with deeper pipelining, but is also clamped by the slowest pipeline stage.
In order to increase clock rate while balancing each pipeline stage, we implement two pipeline stages within each DNN layer i.e., registers are inserted between addition units and activation units as shown in Figure \ref{fig:pipeline}. 

In conventional CMOS circuits performing binary computing, a higher data precision will slow down the clock rate. However, in SC circuits the clock rate is now independent of data precision. In SC, a higher data precision is achieved by longer bit-streams, while the clock rate should be set to cover the operations in each pipeline stage on just \textbf{1-bit} of data. To measure the performance of the SC pipelined architecture, we define \emph{delay} as the bit-stream length times the clock cycle. In this way, the inverse of the delay is equivalent to the throughput of the pipelined architecture of the SC-based hardware DNN.


\section{Experimental Results}


This section demonstrates the effectiveness of our optimized hardware implementation. We adopted one DRL network for the residential smart grid with one 26-neuron input layer, one 30-neuron hidden layer and one single-neuron output layer to implement the hardware application. Therefore the input layer is consisted of 30 XNOR gates for processing the inputs and weight, 30 26-input APCs and Btanh as the activation function. The hidden layer mainly includes a 30-input APC. Converters between stochastic and binary numbers are employed when processing the inputs and generating the outputs. Table \ref{table:binary} presents the hardware implementation of the fixed network using conventional binary computing with the bit size ranging from 8 bits to 32 bits. It can be observed that the SC-based implementation can achieve a much smaller power and area cost compared with the binary-based hardware implementations. 
Table \ref{table: result} shows the result of our proposed DRL hardware implementation based on SC with the impact of pipelining. The bit stream length ranges from 256 to 1024. As showed in the table, the pipelined optimization can significantly reduce the delay, i.e. increase the system throughput, while maintaining small power and area cost. 

\begin{table}[t]
\centering
\caption{Performance of Binary-based Hardware Implementation of the DRL Framework}
\label{table:binary}
\renewcommand{\arraystretch}{1.4}
\begin{tabular}{|c|ccc|}
\hline
\multirow{2}{*}{\textbf{Bit Size}} & \multicolumn{3}{c|}{\textbf{Performance}} \\ \cline{2-4}
                                                                             & \textbf{Delay($ns$)}     & \textbf{Power($mW$)}     & \textbf{Area($\mu m^{2}$)}    \\ \hline
8                                                                          &   7.60        & 63.31          &  1056958.13       \\
16                                                                         &    10.53       &  217.79          & 1080106.41        \\
32                                                                          &  14.76         &   880.25        & 3450187.80      \\ \hline
\end{tabular}
\end{table}

\begin{table}[t]
\centering
\caption{Performance of Optimized Hardware Implementation of the DRL Framework}
\label{table: result}
\begin{adjustbox}{max width=\textwidth,center}
\begin{tabular}{|c|cccc|}
\hline
\multirow{2}{*}{\begin{tabular}[c]{@{}l@{}}\textbf{Bit Stream}\\ \textbf{Length}\end{tabular}} & \multicolumn{4}{c|}{\textbf{Performance}} \\ \cline{2-5} 
 & \multicolumn{1}{c|}{\textbf{Optimization}} & \multicolumn{1}{c|}{\textbf{Delay($ns$)}} & \multicolumn{1}{c|}{\textbf{Power($mW$)} } & \multicolumn{1}{c|}{\textbf{Area($\mu m^{2}$)}} \\ \hline
\multirow{2}{*}{\textbf{256}}  & Pipelining     &261.12  &7.73 &     58771.53                      \\ 
                              & Non-pipelining &   412.47  & 6.30  &  57941.61  \\ \hline
\multirow{2}{*}{\textbf{512}}   & Pipelining     & 522.24 &  7.73   & 58820.74                          \\ 
                               & Non-pipelining    &824.63  &  6.30& 57990.82     \\ \hline
\multirow{2}{*}{\textbf{1024}}  & Pipelining   & 1044.48    &  7.73  &   58919.16                        \\ 
                               & Non-pipelining &  1648.95  &   10.76   & 58089.24    \\ \hline
\end{tabular}

\end{adjustbox}
\end{table}

\section{Conclusion}

In this paper, we first present the general DRL framework, which can be widely utilized in many applications with different optimization objectives. This is followed by the introduction of three specific applications: the cloud computing resource allocation problem, the residential smart grid task scheduling problem, and building HVAC system optimal control problem. The effectiveness of the DRL technique in these three cyber-physical applications have been validated. Finally, this paper investigates the stochastic computing-based hardware implementations of the DRL framework, which consumes a significant improvement in area efficiency and power consumption compared with binary-based implementation counterparts.

\section{Acknowledgements}

This work was supported in part by the National Science Foundation under grants CCF-1553757, CCF-1646381, CNS-1739748 and CNS-1704662, CASE Center at Syracuse University, and Riverside Public Utilities.

\bibliographystyle{IEEEtran}
\bibliography{references}

\end{document}